\documentclass[runningheads,a4paper]{llncs}
\usepackage[T1]{fontenc}
\usepackage{graphicx}
\usepackage{amsmath}
\usepackage{booktabs}
\usepackage{float}
\usepackage{url}
\usepackage{color}
\usepackage[hidelinks]{hyperref}

\begin{document}
\title{EVIL-Detect for NLPCC 2026 Shared Task 6: LLM-Generated Text Detection}
\titlerunning{EVIL-Detect for NLPCC 2026 Task 6}
\author{Hongrui Bao\inst{1,2} \and
Hangyu Rong\inst{2} \and
Zhuoshang Wang\inst{1,2} \and
Yubing Ren\inst{1,2}\thanks{Corresponding author.} \and
Yanan Cao\inst{1,2}}
\authorrunning{H. Bao et al.}
\institute{Institute of Information Engineering, Chinese Academy of Sciences, Beijing, China\\
\email{\{wangzhuoshang,renyubing,caoyanan\}@iie.ac.cn}
\and
School of Cyber Security, University of Chinese Academy of Sciences, Beijing, China\\
\email{\{baohongrui26,ronghangyu23\}@mails.ucas.ac.cn}}

%

%
\maketitle              
\begin{abstract}
The rapid development of large language models (LLMs) has increased the need for reliable detection of LLM-generated text, especially in realistic Chinese scenarios involving human-written text (HWT), LLM-generated text (LGT), and LLM-refined text (HLT). This paper presents EVIL-Detect, a multi-signal ensemble framework with conflict-aware fusion for Natural Language Processing and Chinese Computing (NLPCC) 2026 Shared Task 6. The system integrates edit-extent regression, zero-shot likelihood-contrast signals, lexical statistics, and conservative text rules. With calibrated decision boundaries and conflict-aware integration, our system improves robustness under strong out-of-distribution shifts, achieving a macro-F1 score of 0.8888 and ranking first in the official evaluation. Our code is available at \url{https://github.com/bbbbhrrrr/evildetect}.


\end{abstract}
\section{Introduction}

Large language models (LLMs) such as GPT-4~\cite{ref1}, Qwen2.5~\cite{ref2}, and Qwen3~\cite{ref19} can produce fluent, coherent, and instruction-following text. Their wide use also increases risks such as disinformation, academic misconduct, and manipulation of online content, making reliable LLM-generated text detection important for responsible natural language processing (NLP)~\cite{ref3}.

Most existing work focuses on English binary detection. Supervised detectors, including early GPT-style detectors~\cite{ref4}, RADAR~\cite{ref5}, and DeTeCtive~\cite{ref6}, fine-tune classifiers on labeled data and often perform well in matched settings, but their robustness drops under cross-domain or cross-generator shifts. Training-free methods, including DetectGPT~\cite{ref7}, Fast-DetectGPT~\cite{ref8}, DNA-GPT~\cite{ref9}, and Binoculars~\cite{ref10}, use likelihood-, perturbation-, n-gram-, or perplexity-based signals, but remain sensitive to model choice and thresholds.

Chinese detection is less explored and cannot be handled by direct transfer from English because of differences in tokenization, lexical granularity, and writing conventions. NLPCC 2025 Shared Task 1~\cite{ref11} promoted Chinese binary detection, while NLPCC 2026 Shared Task 6~\cite{ref14} extends the task to three classes: human-written text (HWT), LLM-generated text (LGT), and LLM-refined text (HLT). This setting reflects real artificial intelligence (AI)-assisted writing, where users often use LLMs to rewrite or polish human-written text, and is closely related to CUDRT~\cite{ref12}, DetectRL-X~\cite{ref13}, and the official NLPCC 2026 benchmark~\cite{ref14}.

Motivated by this strong out-of-distribution (OOD) setting, we propose EVIL-Detect, short for Edit-aware View-Integrated Learning for Detection. It combines EditLens-inspired edit-extent regression~\cite{ref15}, zero-shot likelihood-contrast scoring, lexical statistics, and conservative rule-based corrections. Calibrated decision boundaries and conflict-aware integration are used to fuse complementary signals and resolve boundary cases among HWT, LGT, and HLT.

Our contributions are threefold:
\begin{enumerate}
    \item We analyze the performance of several representative detection methods in the Chinese three-class setting of HWT, LGT, and HLT.
    \item We propose EVIL-Detect, a conflict-aware ensemble framework that integrates edit-extent regression, zero-shot likelihood evidence, lexical statistics, and conservative rule-based corrections.
    \item Our system achieved a macro-F1 score of 0.8888 and ranked first in NLPCC 2026 Shared Task 6.
\end{enumerate}

\section{Related Work}

\subsection{Training-based Supervised Detection}

Training-based detectors formulate LLM-generated text detection as supervised classification. Early supervised detectors were developed for GPT-style outputs~\cite{ref4}, and later methods such as RADAR~\cite{ref5} and DeTeCtive~\cite{ref6} improved robustness through adversarial learning and contrastive learning. These methods can be strong when training and test data are similar, but they are vulnerable to shifts in domain, generator, prompt style, and text length~\cite{ref3}.

\subsection{Training-free, Statistical, and Representation-based Detection}

Training-free methods avoid task-specific training and instead rely on language-model statistics. DetectGPT~\cite{ref7}, Fast-DetectGPT~\cite{ref8}, DNA-GPT~\cite{ref9}, Binoculars~\cite{ref10}, and EchoPrompt~\cite{ref25} exploit probability curvature, divergent n-grams, cross-model perplexity, or latent prompt restoration for zero-shot detection. For related fine-grained and robust detection settings, AI-editing-extent regression and representation-level features are also useful: EditLens~\cite{ref15} models the degree of AI editing, while RepreGuard~\cite{ref16} detects generated text from hidden representation patterns.

\subsection{Hybrid Ensemble Detection and Chinese Shared Tasks}

Hybrid and ensemble methods are important because different detectors fail on different samples. EnsemJudge~\cite{ref17}, the winning system of NLPCC 2025 Shared Task 1, combines multiple models and voting strategies for Chinese LLM-generated text detection. CUDRT~\cite{ref12}, DetectRL~\cite{ref18}, DetectRL-X~\cite{ref13}, and NLPCC 2026 Shared Task 6~\cite{ref14} further emphasize realistic, OOD, and AI-edited scenarios. These studies motivate our use of heterogeneous signals and conflict resolution for robust three-class detection.

\section{Method}
\label{sec:method}

In this section, we present EVIL-Detect, our detection system for NLPCC 2026 Shared Task 6. We first describe the observations that motivate the design, then detail the individual modules, and finally present the fusion strategy that combines them into a final prediction.

\subsection{Observation and Motivation}
\label{sec:observation}

Most existing methods for LLM-generated text detection are designed for binary classification, distinguishing human-written text from LLM-generated text. However, NLPCC 2026 Shared Task 6 requires a more fine-grained three-way classification among HWT, LGT, and HLT. A direct adaptation to three-way classification is not sufficiently robust under the distribution shift between training and evaluation data, as later shown by the representative alternative designs in Sec.~\ref{sec:components}. For example, a direct generative quantized low-rank adaptation (QLoRA)~\cite{ref23} classifier achieves only 0.1690 macro-F1 on testp1.

This motivates us to reconsider the label structure of the task. HWT and LGT can be viewed as two endpoints, whereas HLT is an intermediate state because LLM refinement preserves human content while introducing machine-related traces. We therefore do not require every component to solve the full three-way task independently: LGT-oriented signals can be reused by grouping HWT and HLT as non-LGT, while HWT/HLT discrimination relies more on editing strength. Since different methods show different class preferences, as quantified in Sec.~\ref{sec:components}, EVIL-Detect combines a supervised edit-strength module, zero-shot LGT evidence, lexical statistics, and conflict-aware fusion rather than uniform averaging.

\subsection{Overview of EVIL-Detect}
\label{sec:overview}

\begin{figure}[t]
    \centering
    \includegraphics[width=0.9\linewidth]{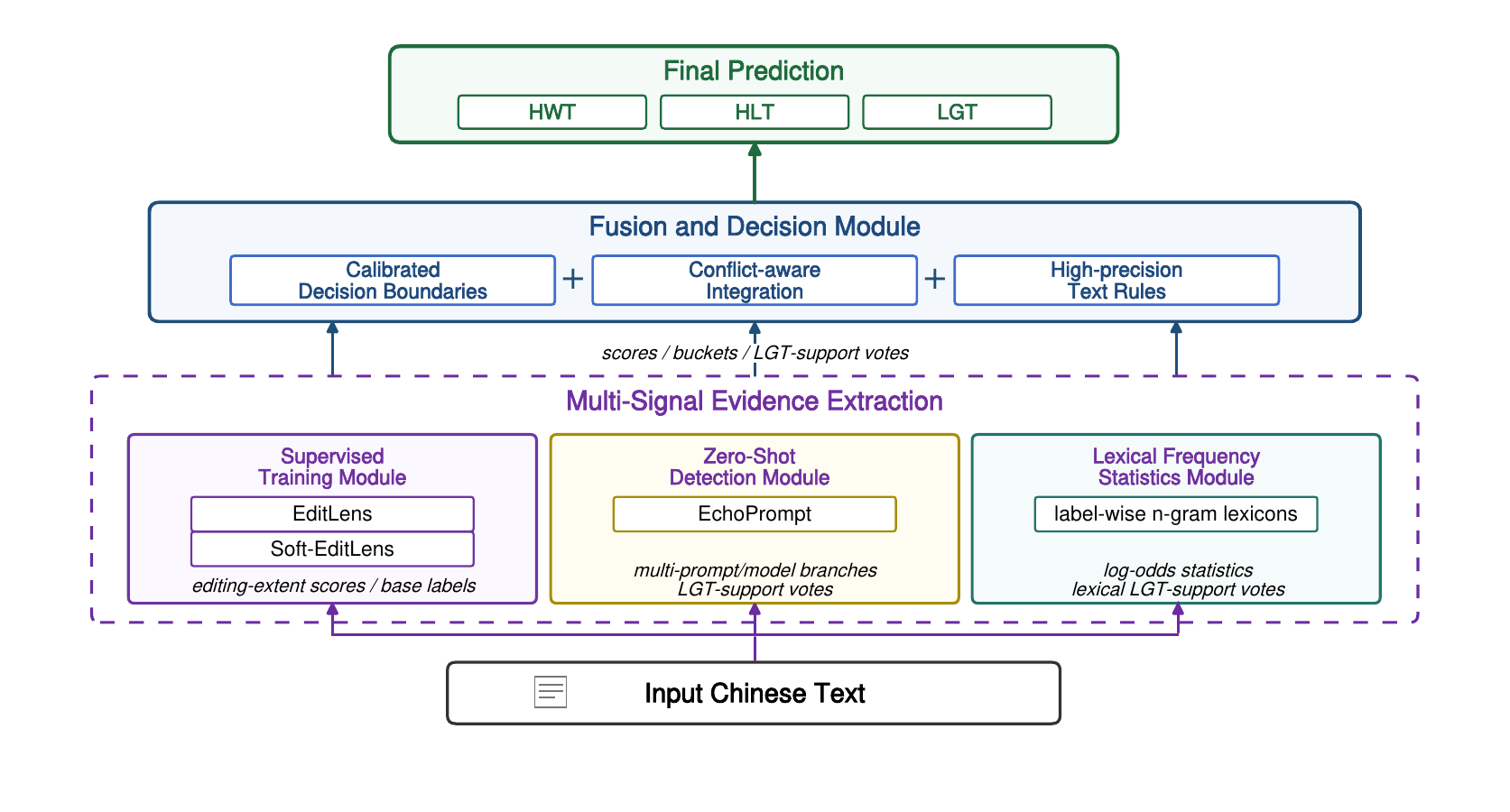}
    \caption{Overall architecture of EVIL-Detect.}
    \label{fig:architecture}
\end{figure}

We propose \textbf{EVIL-Detect}, a multi-signal system for Chinese LLM-generated and LLM-refined text detection. As shown in Fig.~\ref{fig:architecture}, EVIL-Detect combines four types of evidence: supervised edit-extent signals from EditLens~\cite{ref15} and Soft-EditLens, zero-shot LGT-tendency signals from EchoPrompt, lexical frequency statistics, and conservative text rules. The fusion module integrates these heterogeneous signals with conflict-aware decision logic and outputs the final HWT, LGT, or HLT prediction.

\subsection{Supervised Training Module}
\label{sec:supervised}

The supervised training module is inspired by EditLens~\cite{ref15}, which formulates AI-edited text detection as continuous edit-extent estimation rather than pure discrete classification. In the original formulation, a similarity metric between a human-written source and its AI-edited version is used as intermediate supervision, and a regression model is trained to predict the amount of AI editing from the text alone.

In our setting, each training group contains a triplet \((h,g,t)\), where \(h\), \(g\), and \(t\) denote the HWT, LGT, and HLT versions respectively. We define a continuous editing-extent target on the HWT--HLT--LGT axis by using HWT and LGT as two anchors:
\begin{equation}
\label{eq:edit-target}
r(h)=0,\quad r(g)=1,\quad r(t)=d(h,t).
\end{equation}
Here \(d(h,t)\in[0,1]\) measures how far the LLM-refined text \(t\) moves away from the corresponding human-written source \(h\). This formulation treats HLT as an instance-dependent soft editing state rather than an independent hard class. At inference time, the paired HWT text is not required; the trained model predicts \(r(x)\) from the input text \(x\) alone. Based on this common formulation, we instantiate the supervised module with two variants.

\subsubsection{EditLens}
The EditLens variant uses a surface-level soft character \(n\)-gram distance to construct \(d(h,t)\). Let \(C_n(x)\) be the character \(n\)-gram count vector of text \(x\). For a set of \(n\)-gram orders \(\mathcal{N}\) and non-negative weights \(w_n\), we compute
\begin{equation}
\label{eq:ngram-distance}
d_{\mathrm{ng}}(h,t)=1-\frac{\sum_{n\in\mathcal{N}} w_n\,\cos(C_n(h),C_n(t))}{\sum_{n\in\mathcal{N}} w_n}.
\end{equation}
Different \(n\)-gram orders and weights produce different HLT target distributions. We therefore perform a validation-set sweep over candidate \(n\)-gram ranges and weighting schemes, ranking each configuration by how well the resulting HLT scores are separated from the two anchors \(0\) (HWT) and \(1\) (LGT). The best-ranked configuration has internal sweep id 044, uses orders 1--7 with linearly increasing weights, and is denoted as \emph{rank044} in the rest of the paper. The resulting id/text/score instances are used to train a low-rank adaptation (LoRA)~\cite{ref22}-adapted decoder-only LLM with a regression objective. The trained model outputs a continuous editing-extent score, which is later discretized and used as a base signal by the fusion module.

\subsubsection{Soft-EditLens}
Soft-EditLens replaces surface \(n\)-gram overlap with semantic phrase-level soft matching. We first segment HWT and HLT texts into short token phrases. For each phrase \(q\) in the HLT text, we encode it and find the most similar phrase \(p\) in the corresponding HWT text. With phrase count \(c_t(q)\) and embedding function \(e(\cdot)\), the phrase-level distance is
\begin{equation}
\label{eq:phrase-distance}
d_{\mathrm{ph}}(h,t)=1-
\frac{\sum_{q\in P(t)} c_t(q)\max_{p\in P(h)}\cos(e(q),e(p))}{\sum_{q\in P(t)} c_t(q)}.
\end{equation}
Compared with character \(n\)-gram overlap, this target is less sensitive to lexical replacement and better captures semantically preserved expressions. Soft-EditLens also changes the learning objective: besides direct regression on the continuous score, we train an ordinal bucket predictor that discretizes the editing-extent axis into ordered intervals. At inference time, the regression model provides a smooth score, while the bucket model provides an interval-level estimate. These two outputs are used as complementary signals in the fusion module, especially for resolving HWT/HLT boundary cases.

\subsection{Zero-Shot Detection Module}
\label{sec:zeroshot}

\subsubsection{EchoPrompt}
EchoPrompt~\cite{ref25} is a training-free prompting module that provides auxiliary evidence for LGT detection. The module is based on the intuition that fully LLM-generated text is more compatible with an assistant-style generation distribution, whereas human-written text is less likely to follow such a distribution. For each input text, EchoPrompt computes a normalized likelihood-contrast score using a prompted instruction-tuned model and its corresponding base model; a higher score indicates stronger LGT tendency. To reduce the dependence on a single prompt or model, we use multiple fixed prompt/model branches, whose outputs are passed to the fusion module as additional discriminative signals rather than standalone predictions.

\subsection{Lexical Frequency Statistics Module}
\label{sec:lexical}

The lexical frequency statistics module provides a lightweight surface-level view complementary to neural and prompting-based signals. It is based on the observation that some class preferences are reflected in recurring lexical or character \(n\)-gram patterns. We build label-wise frequency lexicons from the training set and retain discriminative \(n\)-grams after removing patterns that appear similarly across classes.

For an input text \(x\), let \(G(x)\) denote its extracted \(n\)-grams. We estimate smoothed label-conditional probabilities \(p(g\mid y)\) for each lexical unit \(g\) and label \(y\), and aggregate their preferences through averaged log-odds scores:
\begin{equation}
\label{eq:lexical-logodds}
s_{a/b}(x)=\frac{1}{|G(x)|}\sum_{g\in G(x)}
\log \frac{p(g\mid a)}{p(g\mid b)} .
\end{equation}
In practice, we use several derived statistics, including LGT-vs-HWT tendency, AI-vs-HWT tendency where LGT and HLT are merged as AI-assisted text, the fraction of machine-oriented lexical units, and the fraction of HWT-oriented lexical units. These statistics are passed to the fusion module as auxiliary lexical evidence rather than used as standalone predictions.

\subsection{Fusion and Decision Module}
\label{sec:fusion}

The fusion module integrates the outputs of the preceding components into the final HWT/LGT/HLT prediction. The starting point is the EditLens editing-extent score. We use two calibrated boundary settings \((\tau_1^{(k)},\tau_2^{(k)})\), \(k\in\{1,2\}\), and convert the score into a base label for each setting:
\begin{equation}
\label{eq:base-label}
y^{(k)}(x)=
\begin{cases}
\mathrm{HWT}, & s_E(x) \le \tau_1^{(k)},\\
\mathrm{HLT}, & \tau_1^{(k)} < s_E(x) < \tau_2^{(k)},\\
\mathrm{LGT}, & s_E(x) \ge \tau_2^{(k)},
\end{cases}
\quad k\in\{1,2\}.
\end{equation}
Here \(s_E(x)\) denotes the EditLens score, and \(y^{(1)}\) and \(y^{(2)}\) are the two resulting base labels. These base labels are combined with Soft-EditLens regression/bucket signals and a panel of nine binary LGT-support votes from zero-shot, edit-based, and lexical views (Table~\ref{tab:votes}).

\subsubsection{Conflict-aware integration}
If \(y^{(1)}\) and \(y^{(2)}\) agree, EVIL-Detect directly keeps the agreed label. If they disagree, the fusion module resolves the conflict according to the label pair involved:
\begin{equation}
\label{eq:base-resolver}
y_{\mathrm{base}} =
\begin{cases}
y^{(1)}, & y^{(1)}=y^{(2)},\\
R(y^{(1)},y^{(2)}, z_{\mathrm{soft}}, V_{\mathrm{LGT}}), & y^{(1)}\ne y^{(2)},
\end{cases}
\end{equation}
where \(z_{\mathrm{soft}}\) represents the Soft-EditLens regression and bucket evidence, \(V_{\mathrm{LGT}}\) denotes the aggregated LGT-support votes, and \(R(\cdot)\) is the conflict-aware resolver.

We instantiate \(R(\cdot)\) with validation-calibrated thresholds. Let \(z_r(x)\) be the Soft-EditLens regression score, \(z_b(x)\) be its ordinal bucket index, and \(v(x)=\sum_{i=1}^{9} v_i(x)\) be the number of active LGT-support votes. For HWT/LGT conflicts, \(R\) outputs HWT when \(z_r(x)\le\gamma_H\), otherwise outputs LGT when \(v(x)\ge\gamma_{HL}\), and otherwise falls back to HLT. For HWT/HLT conflicts, \(R\) outputs HLT iff \(z_b(x)\ge\beta_T\); otherwise it outputs HWT. For HLT/LGT conflicts, \(R\) outputs LGT iff \(v(x)\ge\gamma_{TL}\); otherwise it keeps HLT. Here \(\gamma_H\) is the HWT-side Soft-EditLens boundary, \(\beta_T\) is the rewriting-strength bucket boundary, and \(\gamma_{HL},\gamma_{TL}\) are pair-specific LGT-vote thresholds. This rule design matches the role of each module: EditLens gives the initial continuum label, Soft-EditLens refines HWT/HLT boundaries, and the vote panel provides evidence for fully generated text.

\subsubsection{High-precision text rules}
After conflict-aware integration, we apply a small set of conservative text rules as final corrections. The complete rule set is as follows. First, raw HyperText Markup Language (HTML)-like or Extensible Markup Language (XML)-like structural markup, such as \texttt{<!DOCTYPE html>}, \texttt{<html>}, \texttt{<div>}, \texttt{<p>}, and \texttt{<a href=...>}, is treated as strong evidence for LGT; representative training examples are shown in Appendix~\ref{app:html}. Second, script or rendering residues, such as \texttt{document.write("<br/>")} or unfinished tag sequences, also trigger an LGT correction. Third, explicit rewriting or polishing traces, such as answer wrappers indicating that a rewritten or polished version follows, support HLT when the fused label is not already LGT. These rules are applied only after fusion and serve as high-precision safeguards.

\section{Experiments}
\label{sec:experiments}

In this section, we evaluate EVIL-Detect on NLPCC 2026 Shared Task 6.

\subsection{Experimental Setup}
\label{sec:setup}

\subsubsection{Dataset}
We use the official data released for NLPCC 2026 Shared Task 6, which requires three-way classification among human-written text (HWT), LLM-generated text (LGT), and LLM-refined text (HLT). According to the official task description~\cite{ref14}, the training set is sampled and adapted from the CUDRT dataset~\cite{ref12} and covers four generators, including GPT-4~\cite{ref1}, Qwen-family generators, ChatGLM~\cite{ref20}, and Baichuan~\cite{ref21}, across two domains, namely news and academic writing. In contrast, the two hidden test phases, testp1 and testp2, are drawn from the Chinese split of DetectRL-X~\cite{ref13} and are constructed as out-of-distribution data that may involve unseen domains, unseen generators, and different generation schemes, so as to stress-test detector robustness. We remove empty or malformed training samples and clean task-irrelevant artifacts such as role markers, abnormal control characters, redundant whitespace, and obvious generation residues. Formatting cues such as Markdown or HTML fragments are handled conservatively to avoid removing genuine stylistic signals. This preprocessing is applied only to the training data. Detailed data statistics are shown in Table~\ref{tab:data}.

\begin{table}[t]
\centering
\caption{Dataset statistics.}\label{tab:data}
\begin{tabular}{lcccc}
\toprule
Split & HWT & LGT & HLT & Total\\
\midrule
Cleaned training set & 19{,}634 & 18{,}368 & 19{,}587 & 57{,}589\\
testp1 (official) & 1{,}200 & 1{,}200 & 1{,}200 & 3{,}600\\
testp2 (official) & 384 & 384 & 384 & 1{,}152\\
\bottomrule
\end{tabular}
\end{table}

\subsubsection{Metric.} The official metric is macro-F1 over the three classes. We additionally report accuracy and per-class F1.

\subsubsection{Implementation.} Table~\ref{tab:modules} summarises the modules, and Table~\ref{tab:votes} details the vote panel. EditLens uses Qwen3.5-4B-Base~\cite{ref24} as the backbone with LoRA~\cite{ref22} (rank 16, $\alpha$ 32, dropout 0.05) on all attention and MLP projections, together with a linear regression head for editing-extent prediction. Its target adopts the rank044 soft $n$-gram configuration selected by the separability sweep. Soft-EditLens uses Qwen3.5-9B-Base~\cite{ref24} with the same LoRA~\cite{ref22} configuration and a dual regression/bucket objective. The zero-shot module is training-free and uses Qwen3.5-4B~\cite{ref24} and Qwen2.5-1.5B~\cite{ref2} instruct/base model pairs. The lexical-frequency module builds label-wise word and character $n$-gram log-odds lexicons. For fusion, EditLens scores are discretized with two calibrated class-boundary settings and combined with Soft-EditLens auxiliary signals and nine binary LGT-support votes through conflict-aware integration, followed by high-precision text rules. All supervised models are trained on NVIDIA V100 graphics processing units (GPUs).

\begin{table}[t]
\centering
\caption{Module configuration of EVIL-Detect.}\label{tab:modules}
\begin{tabular}{llll}
\toprule
Module & Backbone & Role & Type\\
\midrule
EditLens & Qwen3.5-4B-Base + LoRA & editing-extent label & supervised\\
Soft-EditLens & Qwen3.5-9B-Base + LoRA & regression/bucket scores & supervised\\
EchoPrompt & Qwen3.5-4B, Qwen2.5-1.5B & LGT vote panel & training-free\\
Lexical freq. & label-wise $n$-gram lexicon & LGT vote panel & training-free\\
Fusion & --- & evidence integration & deterministic\\
\bottomrule
\end{tabular}
\end{table}

\begin{table}[t]
\centering
\caption{Composition of the binary LGT-support vote panel used by the fusion module.}\label{tab:votes}
\begin{tabular}{lll}
\toprule
\# & Source view & LGT-support signal\\
\midrule
1 & EchoPrompt & likelihood-contrast vote (Qwen2.5-1.5B, prefix 1)\\
2 & EchoPrompt & likelihood-contrast vote (Qwen3.5-4B, prefix 1)\\
3 & EchoPrompt & likelihood-contrast vote (Qwen3.5-4B, prefix 2)\\
4 & EditLens & primary editing-extent LGT vote\\
5 & EditLens & auxiliary editing-extent LGT vote\\
6 & Lexical & low HWT-oriented lexical evidence\\
7 & Lexical & AI-vs-HWT lexical log-odds\\
8 & Lexical & LGT-vs-HWT lexical log-odds\\
9 & Lexical & machine-oriented lexical fraction\\
\bottomrule
\end{tabular}
\end{table}

\subsection{Main Results}
\label{sec:mainresults}
Table~\ref{tab:main} reports the performance of EVIL-Detect on the two official test phases. The system achieves macro-F1 scores of 0.8913 on testp1 and 0.8888 on testp2. The small gap between the two phases indicates that the multi-signal design remains stable across the two evaluation splits.

Among the three classes, LGT obtains the highest F1 scores, reaching 0.9267 on testp1 and 0.9219 on testp2. HLT is consistently the most challenging class, with F1 scores of 0.8391 and 0.8407, which is consistent with its intermediate nature between HWT and LGT. This pattern supports the need for conflict-aware fusion, where HWT/HLT and HLT/LGT disagreements are handled with different auxiliary signals.

\begin{table}[t]
\centering
\caption{Main results of EVIL-Detect on the two official test phases of NLPCC 2026 Task 6.}\label{tab:main}
\begin{tabular}{lcccccc}
\toprule
Phase & \#Samples & Macro-F1 & Accuracy & HWT-F1 & LGT-F1 & HLT-F1\\
\midrule
testp1 & 3{,}600 & 0.8913 & 0.8911 & 0.9083 & 0.9267 & 0.8391\\
testp2 & 1{,}152 & 0.8888 & 0.8880 & 0.9039 & 0.9219 & 0.8407\\
\bottomrule
\end{tabular}
\end{table}

\subsection{Component and Variant Analysis}
\label{sec:components}

We evaluate each module in isolation on testp1, together with the main variants explored during development (Table~\ref{tab:components}). EditLens~\cite{ref15} is the strongest standalone component: the sweep-selected rank044 configuration reaches 0.8494 macro-F1, outperforming the initial soft-\(n\)-gram configuration (0.8289). This confirms that the validation-based target sweep improves the separability of the HWT--HLT--LGT continuum.

The other modules are weaker as standalone three-class predictors, but they provide useful auxiliary evidence for fusion. The best zero-shot strategy reaches 0.6325 macro-F1, while the binary HWT/AI strategy reaches 0.4484, indicating that EchoPrompt is better suited for producing LGT-support votes. Lexical statistics show a similar pattern: the best lexical classifier reaches 0.5535 macro-F1, so lexical features are also used as calibrated votes rather than final labels. Soft-EditLens is not listed because its regression and bucket outputs are consumed by the fusion module. With conflict-aware fusion, EVIL-Detect reaches 0.8913 macro-F1, improving over EditLens by 4.19 points and raising HLT and LGT F1 to 0.8391 and 0.9267, respectively.

\begin{table}[t]
\centering
\caption{Component analysis on testp1. Standalone scores are diagnostic because some methods are used as auxiliary evidence sources in the final fusion.}
\label{tab:components}
\begin{tabular}{lcccc}
\toprule
Method & Macro-F1 & HWT & LGT & HLT\\
\midrule
EditLens (rank044) & 0.8494 & 0.9067 & 0.8651 & 0.7765\\
EditLens (initial soft-\(n\)-gram) & 0.8289 & 0.8325 & 0.9042 & 0.7500\\
Zero-shot detection (best strategy) & 0.6325 & 0.6425 & 0.7842 & 0.4708\\
Zero-shot detection (binary HWT/AI) & 0.4484 & 0.6294 & 0.7157 & 0.0000\\
Lexical statistics (best classifier) & 0.5535 & 0.6591 & 0.5040 & 0.4975\\
\midrule
EVIL-Detect (full fusion) & 0.8913 & 0.9083 & 0.9267 & 0.8391\\
\bottomrule
\end{tabular}
\end{table}

\subsubsection{Representative alternative designs.}
Table~\ref{tab:alternatives} reports representative alternatives explored during development but not included in the final EVIL-Detect system. Direct generative supervised fine-tuning (SFT) classifiers are not robust under the evaluation distribution: the GLM-4-9B~\cite{ref20} and Qwen3.5-9B~\cite{ref24} variants reach only 0.1896 and 0.1690 macro-F1, showing strong class bias. Binoculars~\cite{ref10} also performs poorly as a standalone three-way predictor, with 0.3928 macro-F1. These results support our design choice of using editing-extent modeling as the base signal and treating likelihood-based detectors as auxiliary evidence.

\begin{table}[t]
\centering
\caption{Representative alternative designs evaluated on testp1.}
\label{tab:alternatives}
\begin{tabular}{lcccc}
\toprule
Alternative design & Macro-F1 & HWT & LGT & HLT\\
\midrule
Binoculars & 0.3928 & 0.4738 & 0.5552 & 0.1493\\
Generative SFT (GLM-4-9B) & 0.1896 & 0.5082 & 0.0148 & 0.0458\\
Generative SFT (Qwen3.5-9B) & 0.1690 & 0.0000 & 0.0066 & 0.5004\\
\bottomrule
\end{tabular}
\end{table}

\subsection{Ablation Study}
\label{sec:ablation}

We conduct an ablation of the fusion module on the released testp2 labels, as shown in Table~\ref{tab:ablation}. The EditLens-only baseline~\cite{ref15} obtains 0.8411 macro-F1. Adding conflict-aware integration and high-precision text rules improves the score to 0.8816, and the full system reaches 0.8888 after combining two calibrated EditLens decisions. The improvement mainly comes from boundary cases: compared with EditLens only, the full system reduces LGT$\rightarrow$HLT errors from 59 to 30 and HLT$\rightarrow$LGT errors from 58 to 28, showing that the fusion module helps correct confusion around HLT.

\begin{table}[H]
\centering
\caption{Ablation results of the fusion module on testp2.}
\label{tab:ablation}
\begin{tabular}{lcccc}
\toprule
Configuration & Macro-F1 & HWT & LGT & HLT\\
\midrule
Full EVIL-Detect & 0.8888 & 0.9039 & 0.9219 & 0.8407\\
EVIL-Detect (single-threshold EditLens) & 0.8816 & 0.9153 & 0.9016 & 0.8280\\
EditLens only & 0.8411 & 0.9141 & 0.8453 & 0.7640\\
\bottomrule
\end{tabular}
\end{table}

\section{Conclusion}

This paper presents EVIL-Detect for Chinese LLM-generated and LLM-refined text detection in NLPCC 2026 Shared Task 6. It integrates edit-extent regression, zero-shot LGT evidence, lexical statistics, and conservative text rules through conflict-aware fusion. With calibrated decision boundaries, our system achieved a macro-F1 score of 0.8888 and ranked first in the official evaluation. The results suggest that robust three-class detection cannot rely on a single signal; instead, edit-strength modeling, likelihood-based evidence, lexical cues, and high-precision corrections provide complementary information for distinguishing HWT, LGT, and HLT.

\subsubsection*{Acknowledgments.}
This work is supported by the Postdoctoral Fellowship Program of China Postdoctoral Science Foundation (CPSF) under Grant Number GZC20251076.

%
%
%
%

\appendix
\section{HTML-like LGT Examples}
\label{app:html}

We observed LGT training samples with raw markup or rendering residues. Representative short snippets include:
\begin{itemize}
\item \texttt{<!DOCTYPE html>}
\item \texttt{<html>} and \texttt{<div>}
\item \texttt{<a href=...>}
\item \texttt{document.write("<br/>")}
\end{itemize}
These patterns appear in LGT training items generated by Baichuan, ChatGLM, and GPT-4. We show only structural snippets because the original samples are long Chinese documents.

\end{document}